%% file: main.tex
\pdfoutput=1
\documentclass[11pt]{article}
\usepackage[final]{ACL2023}

\usepackage{times}
\usepackage{latexsym}
\usepackage[T1]{fontenc}
\usepackage[utf8]{inputenc}
\usepackage{microtype}
\usepackage{inconsolata}
\usepackage{amsmath}
\usepackage{amsthm} 
\usepackage{graphicx}
\usepackage{subfigure}
\usepackage{booktabs}
\usepackage{xcolor}
\usepackage{listings}
\usepackage{tabularx}
\usepackage{array}
\usepackage{makecell}
\usepackage{multirow}
\usepackage{enumitem}
\usepackage{algorithm}
\usepackage{algorithmic}
\usepackage{amssymb}
\usepackage{url}

\newenvironment{denseitemize}{
\begin{itemize}[topsep=2.5pt, partopsep=0pt, leftmargin=1.5em]
 \setlength{\itemsep}{2.5pt}
 \setlength{\parskip}{0pt}
 \setlength{\parsep}{0pt}
}{\end{itemize}}

\newcommand{\disco}{{\emph{DiSCo}}}

\title{\disco: Device-Server Collaborative LLM-Based Text Streaming Services}

\author{Ting Sun$^{1}$,
Penghan Wang$^{2}$, 
Fan Lai$^{1}$\thanks{\hspace{.1cm} Corresponding author} \\
\textsuperscript{1} University of Illinois Urbana-Champaign, United States \\
\textsuperscript{2} Purdue University, United States \\
\texttt{suntcrick@gmail.com, wangpenghan381@gmail.com, fanlai@illinois.edu} \\
}

\begin{document}
\maketitle

\input{src/abstract.tex}
\input{src/introduction}
\input{src/background.tex}
\input{src/characterization.tex}
\input{src/disco_policy.tex}
\input{src/evaluation.tex}
\input{src/conclusion.tex}

\section{Limitations}
While \disco{} demonstrates significant improvements in LLM serving efficiency, we acknowledge several important limitations of our current work:

\paragraph{Model Accuracy.} We focus on scenarios where on-device LLMs achieve sufficient accuracy for target applications. While this covers many common use cases, \disco{} may not be suitable for complex reasoning tasks~\cite{guo2025deepseek}.

\paragraph{Privacy Protection.} While privacy protection is a key consideration in model selection~\cite{chen2024llm,liu2024shield}, \disco{} currently assumes users are comfortable with both on-device and on-server deployments.

\paragraph{Energy Modeling.} For device energy consumption, we use a linear energy model based on FLOPs. Real-world device energy consumption patterns can be more complicated, varying on factors such as battery state, temperature, and concurrent workloads~\cite{hoque2015modeling}.

\paragraph{Scalability.} Our current implementation and evaluation focus on single-device scenarios. Extending \disco{} to handle multi-device collaborative serving presents additional challenges in terms of coordination overhead and resource allocation that warrant further investigation~\cite{niu2025collaborative}.

\section{Ethical Considerations}
Our work focuses solely on optimizing the efficiency of LLM serving systems through device-server collaboration and does not introduce new language generation capabilities or content. All experiments were conducted using publicly available models and datasets. While our work may indirectly benefit the accessibility of LLM services by reducing costs and improving performance, we acknowledge that broader ethical considerations around LLM deployment and usage are important but outside the scope of this technical contribution.

\bibliography{main}
\bibliographystyle{acl_natbib}

\appendix
\input{src/appendix/related_work.tex}
\input{src/appendix/cold_start.tex}
\input{src/appendix/prediction.tex}
\input{src/appendix/response_quality.tex}
\input{src/appendix/unified_cost.tex}
\input{src/appendix/pseudocode.tex}

\end{document}

%% file: src/abstract.tex
\begin{abstract}
The rapid rise of large language models (LLMs) in text streaming services has introduced significant cost and Quality of Experience (QoE) challenges in serving millions of daily requests, especially in meeting Time-To-First-Token (TTFT) and Time-Between-Token (TBT) requirements for real-time interactions. Our real-world measurements show that both server-based and on-device deployments struggle to meet diverse QoE demands: server deployments face high costs and last-hop issues (e.g., Internet latency and dynamics), while on-device LLM inference is constrained by resources.
    
We introduce \disco{}, a device-server cooperative scheduler designed to optimize users' QoE by adaptively routing requests and migrating response generation between endpoints while maintaining cost constraints. \disco{} employs cost-aware scheduling, leveraging the predictable speed of on-device LLM inference with the flexible capacity of server-based inference to dispatch requests on the fly, while introducing a token-level migration mechanism to ensure consistent token delivery during migration. Evaluations on real-world workloads---including commercial services like OpenAI GPT and DeepSeek, and open-source deployments such as LLaMA3---show that \disco{} can improve users' QoE by reducing tail TTFT (11-52\%) and mean TTFT (6-78\%) across different model-device configurations, while dramatically reducing serving costs by up to 84\% through its migration mechanism while maintaining comparable QoE levels.
\end{abstract}

%% file: src/introduction.tex
\section{Introduction}

Large language models (LLMs) have revolutionized various applications, with over 60\% focusing on conversational interactions such as chatbots \citep{llm-market-report}. 
Meeting high serving demands requires scaling deployments across on-premise servers in the cloud and on-device inference, as seen in Apple Intelligence \citep{appleintelligence} and Google's Gemini Nano \citep{gemini_nano}.
The Quality of Experience (QoE) for interactive applications is primarily evaluated by two critical metrics: Time-To-First-Token (TTFT) in the prefill stage, which quantifies the initial response latency, and Time-Between-Token (TBT) during the decode stage, which measures the consistency of token delivery speed~\citep{databricks2023llm,andes,cachegen}.

On-server deployments lower serving costs by sharing infrastructure among many requests but often introduce unpredictable high latency due to request queuing delays~\citep{sarathi} and the internet speed fluctuations. While on-device deployment is able to serve increasingly capable LLMs with sufficient accuracy, it suffers from slow processing speeds for long prompts and high energy consumption. For example, a fully-charged iPhone can only operate for less than two hours running an LLM with 7B parameters~\citep{liu2024mobilellm}.

This paper introduces a novel paradigm for cost-constrained device-server cooperative inference. We incorporate both server usage (e.g., monetary costs) and device energy costs via a dynamic exchange rate, which can be adjusted by endpoint users to balance response generation between the cloud and devices. 
As such, we can strategically distribute inference requests between endpoints and dynamically migrate ongoing token generation to maximize QoE. 
However, realizing this vision presents several fundamental challenges: 

\begin{denseitemize}
   \item \textbf{Unified Cost Management:} 
   Total serving costs include resource expenditures from both endpoints---monetary costs from server API usage and energy costs from device computation. The relative value of energy costs varies dynamically based on device context (e.g., battery level, charging status) and user preferences for server spending, making it challenging to establish a unified optimization strategy. 
   
   \item \textbf{Runtime Uncertainty:} 
   The dynamic nature of networks (e.g., latency jitters) and serving loads make it challenging to accurately predict TTFT for in-flight request migration. Moreover, any scheduling mechanism must be lightweight to avoid introducing large overhead to the already latency-sensitive services.
   
   \item \textbf{Migration Impact on Token Delivery:}
   While dynamic migration between endpoints can reduce overall operating costs, it risks disrupting TBT. The challenge lies in determining when and how to migrate while minimizing user experience degradation and cost increase.
\end{denseitemize}

\begin{figure}[t]
   \centering
   \includegraphics[width=0.4\textwidth]{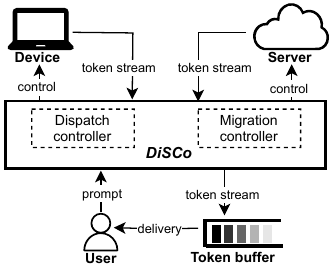}
   \caption{\disco{} acts as a middleware to optimize QoE by adaptively dispatching and migrating response generation between device and server endpoints under cost constraints. 
   }
   \vspace{-.3cm}
   \label{fig:overview}
\end{figure}

As shown in Figure~\ref{fig:overview}, we introduce \disco{}, a \underline{D}ev\underline{i}ce-\underline{S}erver \underline{Co}operative scheduler that addresses these challenges via two key innovations:

\begin{denseitemize}
   \item \textbf{Cost-Aware Dispatching Policies:} 
   We introduce two dispatching mechanisms targeting different cost constraints. For server cost constraints, we employ a length-threshold based dispatching mechanism that routes requests shorter than a dynamically computed threshold to devices. For device energy constraints, we implement a delay-based dispatching mechanism where devices wait for a computed interval before starting local inference. Both mechanisms adapt their thresholds based on unified cost measures that combine server monetary costs and device energy consumption. 

   \item \textbf{Token-Level Migration Framework:} 
   We enable seamless generation handoff between endpoints through a novel migration protocol that preserves the consistency of token delivery. Our framework employs delayed migration timing to minimize interruption, while a token buffer ensures smooth delivery during transitions. This design maintains user experience while saving resource costs across endpoints.
\end{denseitemize}

Through extensive evaluation using real-world traces from commercial LLM streaming API services, including GPT and DeepSeek, and on-device deployments, we demonstrate that \disco{} improves mean and tail TTFT by up to 50\% without TBT violation, significantly reducing costs.

Overall, we make the following contributions:
\begin{denseitemize}
   \item We characterize QoE challenges in device-server cooperative LLM inference through extensive real-world measurements. 
   
   \item We design novel scheduling policies that optimize QoE under cost constraints. 
   
   \item We develop a token-level migration framework to enable generation handoff between endpoints, preserving token delivery consistency.
   
   \item We demonstrate \disco{}'s effectiveness in commercial services and open-source benchmarks.
\end{denseitemize}

%% file: src/background.tex
\section{Background and Motivation}

\subsection{LLM Token Mixture and Routing}
\label{sec:llm_routing}
Device-server collaborative approaches have evolved in two directions. First, systems like EdgeShard~\cite{zhang2024edgeshard} and WDMoE~\citep{xue2024wdmoe} partition LLMs across multiple endpoints when a single device cannot host the entire model. LLMCad~\cite{xu2023llmcad} uses on-device models to reduce server costs, while PerLLM~\citep{yang2024perllm} optimizes energy consumption across devices and servers under constraints. Second, routing-based approaches~\citep{routellm,hybridllm} balance cost and accuracy by directing simple requests to small models and complex queries to advanced ones. However, these approaches do not optimize token delivery metrics (TTFT and TBT) under cost constraints.

\subsection{LLM-Based Text Streaming Applications}
\label{sec:llm_applications}
Over 60\% of LLM-backed applications focus on streaming conversational interactions, such as chatbots, virtual assistants, and language translation.
QoE in these text streaming services is often quantified by two critical metrics: time-to-first-token (TTFT) for \emph{initial responsiveness} and time-between-tokens (TBT) for \emph{delivery smoothness} throughout the entire interaction timeline.

Current LLM systems struggle to meet user expectations for these metrics, with TTFTs ranging from hundreds of milliseconds to over ten seconds—far exceeding the ideal latencies of tens of milliseconds for interactive applications~\cite{maki2004latency,latency_comp_surv}. Token consumption patterns vary by output modality: In visual text scenarios, reading speeds differ across demographic groups, with the majority (52\%) aged 25-44 reading 4-5 tokens per second, while older groups generally read more slowly~\cite{andes,andes_read,word_to_token}. Audio output consumption shows more consistency, averaging 3-4 tokens per second across languages~\cite{andes,andes_speak,average-speaking-rate}. Notably, conventional evaluation metrics like token generation throughput or average time-per-output-token provide incomplete insights, as they fail to capture the crucial relationship between token delivery timing and actual user consumption patterns.

\subsection{Limitations of Existing Text Streaming Applications}
\label{sec:existing_limitations}
Existing LLM serving primarily relies on two deployment paradigms: on-device and on-server inference. With rapid hardware and software advancements, on-device LLMs have achieved sufficient accuracy levels for many applications~\citep{phan2025humanity}, as evidenced by the integration of Apple Intelligence~\citep{appleintelligence} and Google's Gemini Nano~\cite{gemini_nano} into iOS and Android platforms, where they effectively handle text completion and message composition tasks. While on-device LLMs may still be inadequate for complex tasks (e.g., advanced mathematical reasoning), we focus on the growing category of applications where current on-device models already achieve satisfactory accuracy. For these applications, the challenge is not model capability, but rather the substantial monetary or energy cost demands of LLM inference.

Unfortunately, both serving paradigms face challenges. On-device inference, though enabling faster generation powered by its dedicated resources~\citep{powerinfer,powerinfer2}, suffers from extended TTFT for long prompts due to limited processing speeds and substantial energy consumption that scales linearly with response lengths \citep{edgebenchmark}. For instance, a fully-charged iPhone running a 7B parameter LLM can only operate for less than two hours \citep{liu2024mobilellm}---insufficient for day-long mobile use.

On the other hand, on-server deployments require request batching to amortize costs due to the high resource demands, but this introduces issues like queuing delays, resource contention from batching \citep{orca,vllm,sarathi}, and last-hop network latency variations \cite{eloquent}. Our measurements reveal that these factors can cause significant TTFT spikes for GPT-4-mini, from 0.3 seconds to several seconds during high-load periods.

Given these complementary limitations, we investigate the following research question: \emph{Can a cooperative paradigm be designed to combine on-server and on-device inference to improve QoE while managing both energy and monetary costs?} 

%% file: src/characterization.tex
\section{Characterizing LLM Inference}
\label{sec:characteristics}
This section characterizes the LLM inference performance of on-server and on-device paradigms, which informs our design.

We evaluate four commercial streaming LLM APIs: OpenAI's GPT-4o-mini~\citep{gpt-4o-mini}, DeepSeek's DeepSeek-V2.5~\citep{deepseek-v2_5}, Cohere's Command~\citep{command}, and Hyperbolic-hosted LLaMA-3-70b-Instruct~\citep{llama3-70b}. For on-device analysis, we deploy Qwen-2.5-7B-Instruct~\citep{qwen2_5} and Llama-3.1-8B-Instruct~\citep{grattafiori2024llama3herdmodels} on both server-grade (NVIDIA A40, 48GB) and consumer-grade (dual NVIDIA RTX 3080, denoted as 3080x2) GPUs. 
We sample 1,000 requests from the Alpaca dataset~\citep{alpaca}, following a Poisson distribution with a mean request arrival interval of 30 seconds.

\begin{figure}
    \subfigure[On-Server TTFTs.]{\includegraphics[width=0.5\columnwidth]{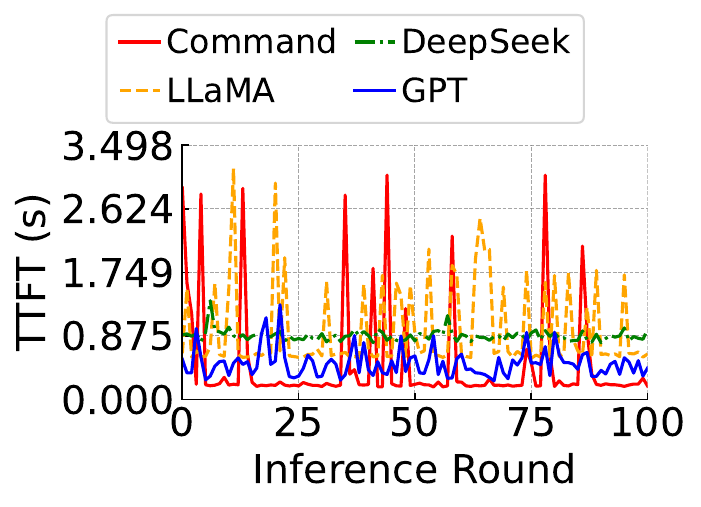}}\hfill
    \subfigure[On-Device TTFTs.]{\includegraphics[width=0.5\columnwidth]{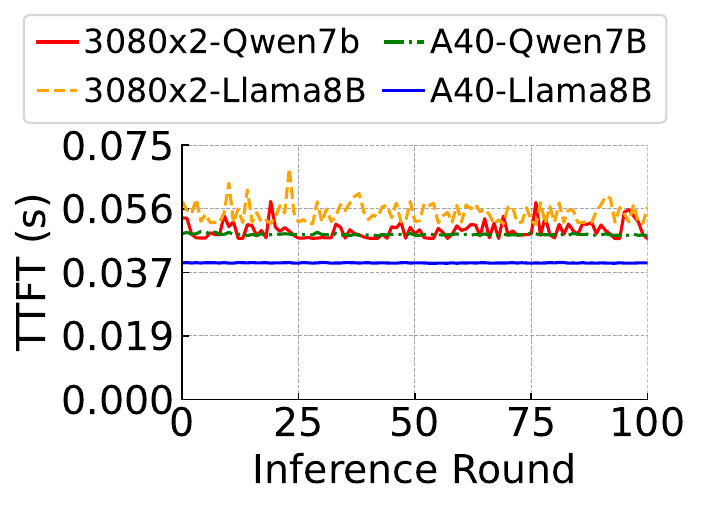}}
    \caption{On-device TTFT performance is more stable.}
    \vspace{-.2cm}
    \label{fig:ttft_repeat}
\end{figure}

\paragraph{TTFT characteristics.} 
Our measurements reveal the contrasting TTFT patterns between on-device and on-server inference. As shown in Figure~\ref{fig:ttft_repeat}, on-device inference exhibits stable TTFTs when processing identical prompts at 60-second intervals, primarily reflecting the prefill duration due to dedicated local hardware resources. In contrast, on-server inference experiences high variations and significant tail latency, attributed to network delays, request queuing, and resource contention.

Unlike previous works that focus solely on prefill latency (e.g., \cite{gim2024prompt,fiddler}) or the sum of queuing and prefill (e.g., \cite{sarathi,mooncake}), we measure user-perceived TTFT as a comprehensive sum of network, queuing, and prefill latencies. Our extensive evaluations across four major LLM providers consistently demonstrate this improvement. This methodology aligns with industry benchmarks, such as Artificial Analysis~\cite{ArtificialAnalysis2025}'s continuous monitoring of diverse LLM services. Notably, since most on-device LLMs struggle with prefilling or understanding long prompts, we focus on short prompts, where request queuing delay and batch completion waiting time during generation dominate the overall TTFT.

We summarize the TTFT performance of 1,000 requests in Table~\ref{tab:correlation-analysis}. We observe that on-device TTFT scales linearly with prompt length due to hardware constraints~\citep{edgebenchmark}, while on-server TTFT shows minimal prompt-length sensitivity through advanced resource scaling~\citep{distserve,splitwise,memserve}. Note that LLM providers (e.g., Microsoft Azure~\cite{Azure2025}, DeepSeek~\cite{DeepSeek2025}, Together.ai~\cite{TogetherAI2025}, Hyperbolic Labs~\cite{Hyperbolic2025}) typically do not offer explicit TTFT SLOs, likely due to the high cost and complexity of maintaining such guarantees across diverse models and prompts.

\begin{table}[t]
    \centering
    \footnotesize
    \begin{tabularx}{\linewidth}{lcc}
    \toprule
    \textbf{Model} & \textbf{Deployment} & \textbf{Pearson Coef.} \\
    \midrule 
    Command & Server & 0.0142 \\
    GPT-4o-mini & Server & 0.0236 \\
    DeepSeek-V2.5 & Server & -0.0273 \\
    LLaMA-3-70b-Instruct & Server & 0.0402 \\
    \hline
    LLaMA-3.1-8b-Instruct & Device & 0.8424 \\
    \bottomrule
    \end{tabularx}
    \caption{Pearson coefficient between prompt length and TTFT in on-server deployment is weak.}
    \vskip -0.1in
    \label{tab:correlation-analysis}
\end{table}

\begin{figure}[t]
    \centering
    \includegraphics[width=0.49\textwidth]{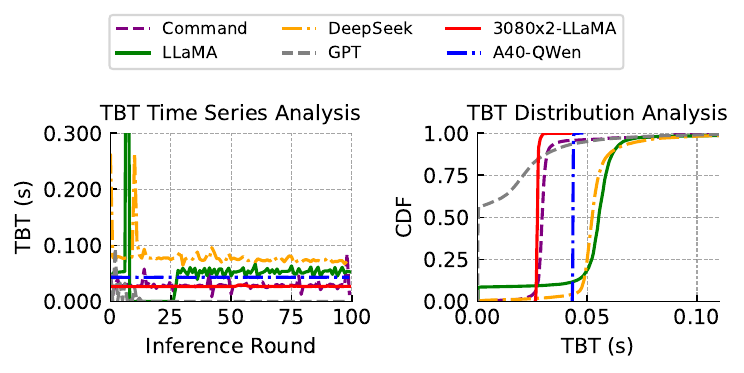}
    \vskip -0.1in
    \caption{On-device TBT performance is more stable. \footnotemark[1]{}
    }
    \vspace{-.3cm}
    \label{fig:tbt_analysis}
\end{figure}

\paragraph{TBT characteristics.}
TBT characterizes the I/O-bound decode stage latency. Analysis of temporal samples and distributions across six setups (Figure~\ref{fig:tbt_analysis}) reveals higher TBT variability in on-server inference compared to on-device execution. Moreover, both deployments achieve generation speeds exceeding user consumption rates~(\S~\ref{sec:llm_applications}), making cooperative serving practical.

\footnotetext[1]{On-server inference, such as in GPT, streams tokens with each packet containing multiple tokens, resulting in near-zero perceived TBTs.}

\paragraph{Opportunities and challenges.}
Our studies further reveal that as on-device models continue to improve---often fine-tuned for specific tasks~\citep{appleintelligence,liu2024mobilellm}---they are progressively achieving performance parity with on-server models in popular applications like instruction-following and translation (detailed in \S\ref{sec:evaluation} and Appendix~\ref{appendix:accuracy-eval}). However, deploying these models on-device introduces challenges such as long prefill latency and startup overhead.

On the other hand, our real-world studies of conversational workloads highlight key opportunities: (i) on-server TTFT is largely unpredictable and shows minimal correlation with prompt length, whereas on-device TTFT scales nearly linearly with prompt length and is highly predictable; and (ii) both paradigms achieve token generation speeds that exceed typical user consumption rates.

By integrating these findings---especially the predictable performance of on-device inference and the elastic scaling capabilities of server-based inference---we observe opportunities for optimization in cost-constrained device-server cooperative serving. Dynamic request migration between server and device endpoints during response generation can yield significant cost savings. 

%% file: src/disco_policy.tex
\section{\disco{} Policies}
\label{sec:disco_policy}

\disco{} optimizes both QoE and cost through (1) dispatch control that determines where to initiate token generation, and (2) migration control that enables dynamic handoff during generation. The dispatch controller optimizes TTFT by strategically routing requests, while the migration controller maintains consistent TBT while reducing costs.

\subsection{Problem Formulation}
\label{subsec:model}
We propose a unified cost model combining both monetary bills from on-server inference and energy bills from on-device inference. Let $c^p_s$ and $c^d_s$ denote the per-token monetary costs for server prefill and decode phases, respectively, while $c^p_d$ and $c^d_d$ represent the per-token energy costs for device prefill and decode phases. Integration of energy and monetary costs is done by a dynamic exchange rate $\lambda$, adjusted by users to reflect their preferences. We offer a user-friendly tunable budget ratio $b \in [0,1]$, representing the additional cost allowance beyond baseline costs. Our optimization objectives focus on: (1) minimizing both mean and tail TTFT, and (2) maintaining consistent token delivery at a specified pace (i.e., stable TBT).

\subsection{Dispatch Controller: Cost-Aware Request Routing}
\label{subsec:ttft_opt}
Based on our analysis in \S\ref{sec:characteristics}, server-side TTFT shows weak correlation with prompt length due to various factors (network delay, request queuing, etc.). We model server TTFT as a known distribution, obtained either from server-provided information or device-side profiling. In contrast, device-side TTFT exhibits a linear relationship with prompt length, with the coefficient determined through offline profiling. 

Our key insight is that the optimization problem naturally decomposes into two scenarios based on dominant cost factors: device-constrained scenarios where energy consumption is the primary bottleneck, and server-constrained scenarios where API monetary costs dominate. This decomposition enables efficient solutions. The pseudocode for the dispatch controller is attached in Appendix~\ref{appendix:pseudocode_scheduling}.

\paragraph{Device-Constrained Optimization.} 

When device costs dominate ($\min(c^p_d, c^d_d) > \max(c^p_s, c^d_s)$), we need to carefully manage device resource usage under a budget constraint $\mathbb{E}[I_d(l)l] \leq b \cdot \mathbb{E}[l]$, where $l$ is the prompt length and $I_d(l)$ indicates device execution. The key challenge is balancing between two goals: leveraging device execution to bound worst-case latency while conserving energy on shorter prompts when possible.

Our solution uses a waiting-time strategy: for each prompt of length $l$, we first try server execution and wait for time $w(l)$ before potentially starting device execution. This conserves device energy when the server responds quickly. We determine the optimal wait time through a two-phase approach:

\begin{denseitemize}
    \item \textbf{Phase 1 (Tail Protection):} We reserve a budget portion $\alpha$ for worst-case scenarios by setting a maximum wait time $w_{tail} = F^{-1}(1-\min(\alpha,b))$, where $F(\cdot)$ is the server TTFT distribution. This ensures we have device resources ready when server latency exceeds its $(1-\min(\alpha,b))$-th percentile.
    
    \item \textbf{Phase 2 (Average Case):} With the remaining budget $(b-\alpha)$, we set length-dependent wait times:
    \begin{equation}
        w(l) = \begin{cases}
            0 & \text{if } l \leq l_{th} \\
            \min(\beta l, w_{tail}) & \text{otherwise}
        \end{cases}
    \end{equation}
    where $l_{th}$ is a threshold below which we start device execution immediately, and $\beta$ is chosen to satisfy:
    \begin{equation}
        \int_{l_{th}}^{\infty} (1-F(\beta l)) \cdot c^p_d \cdot l \cdot p(l)dl = (b-\alpha) \cdot \mathbb{E}[l]
    \end{equation}
\end{denseitemize}

This design guarantees worst-case TTFT through $w_{tail}$ while optimizing average performance by adaptively adjusting wait times based on prompt length. Whichever endpoint (server or device) generates the first token continues to the decode phase, while the other terminates.

\paragraph{Server-Constrained Optimization.}
When server costs dominate ($\max(c^p_s, c^d_s) > \min(c^p_d, c^d_d)$), we need to carefully manage server resource usage under a budget constraint $\mathbb{E}[I_s(l)l] \leq b \cdot \mathbb{E}[l]$, where $I_s(l)$ indicates server execution. Our analysis in \S\ref{sec:characteristics} shows that device TTFT scales linearly with prompt length as $T_d(l) = kl + c$, while server TTFT has minimal length correlation. This suggests a length-based routing strategy: short prompts run on the device to conserve server budget, while long prompts use both endpoints to minimize TTFT.

We determine the length threshold $l_{th}$ by:
\begin{equation}
  \int_0^{l_{th}} l \cdot p(l) dl = (1-b) \cdot \mathbb{E}[l]
\end{equation}
This ensures prompts shorter than $l_{th}$ consume exactly $(1-b)$ fraction of total expected tokens through device-only execution, leaving the remaining longer prompts with sufficient server budget for concurrent execution on both endpoints.

\subsection{Migration Controller: Cost-Efficient Token Delivery}\label{sec:migration}
When both endpoints process a request, the constrained endpoint may win the prefill phase but incur higher decode costs. In such cases, we can migrate token generation to the other endpoint to reduce total cost while maintaining quality.

\paragraph{Theoretical Migration Framework.}
The token-level migration protocol ensures seamless handoffs by leveraging the gap between token generation ($r_g$) and consumption ($r_c$) rates. The migration trigger is determined by comparing the cost savings against migration overhead:
\begin{equation}
    C_{\text{migration}} = \Delta c_{\text{decode}} \cdot l_{\text{remaining}} > \text{Overhead}_{\text{migration}}
\end{equation}
where $\Delta c_{\text{decode}}$ is the per-token cost difference between endpoints and $l_{\text{remaining}}$ represents the expected remaining sequence length. This formulation ensures that migration only occurs when the projected cost savings exceed the overhead of transferring control between endpoints.

\paragraph{Efficient Token Transfer.}

When endpoints share the same vocabulary, we transmit token IDs rather than complete token representations. Empirical analysis using the Alpaca dataset~\cite{alpaca} with the cl100k\_base tokenizer~\cite{tiktoken} (used by GPT-3.5/4) demonstrates that token ID transmission versus UTF-8 encoded text yields 35.62\% reduction in data volume when using minimum byte encoding (3 bytes per token), and 54.40\% reduction in data volume when using minimum bit encoding (17 bits per token). These efficiency gains are particularly valuable in bandwidth-constrained environments, helping to minimize latency during migration.

Additionally, we avoid transferring intermediate states (e.g., attention key-value cache) for two practical reasons: (1) endpoints often employ different model architectures optimized for their respective hardware, making state transfer incompatible, and (2) intermediate state transfer would incur significant network overhead. For models with different vocabularies, we first convert tokens to text before re-tokenizing on the target model to ensure semantic consistency.

Migration is triggered when the projected cost savings exceed overhead:
\begin{equation}
    C_{migration} = \Delta c^d_{decode} \times l_{remaining}
\end{equation}
where $\Delta c^d_{decode} = |c^d_s - c^d_d|$ and $l_{remaining}$ denote the per-token decode cost difference between endpoints, and the expected remaining sequence length, respectively.

\paragraph{Buffer-Based Migration Protocol.}
To ensure smooth token delivery during migration, we introduce a token buffer that leverages the natural gap between token generation speed ($r_g$ tokens/s) and human consumption rate ($r_c$ tokens/s, typically $r_g > r_c$). The buffer size is set to:
\begin{equation}
    B = r_c \times t_m
\end{equation}
where $t_m$ is the estimated migration overhead time. Migration begins only when the buffer contains enough tokens ($B$) to overshadow the migration latency. Importantly, the source endpoint stops generating new tokens once the buffer is filled, preventing any potential conflicts or branching during the transition. This ensures deterministic token delivery as the target endpoint takes over generation.

\begin{figure}[t]
    \centering
    \includegraphics[width=0.45\textwidth]{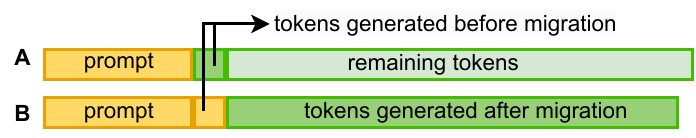}
    \caption{Token generation migration between endpoints. Row A shows the original sequence on the source endpoint, while Row B shows the sequence after migration to the target endpoint, maintaining consistent token delivery while reducing cost.}
    \vspace{-.2cm}
    \label{fig:cost_saving}
\end{figure}

As shown in Figure~\ref{fig:cost_saving}, this design enables seamless handoff: the source endpoint (Row A) continues generation until the buffer is filled, then stops to allow the target endpoint (Row B) to take over, ensuring uninterrupted token delivery to users despite the underlying endpoint transition.

%% file: src/evaluation.tex
\section{Evaluation}
\label{sec:evaluation}

Through extensive experimentation with four production-grade LLM services and state-of-the-art open-source models, we demonstrate \disco{}'s exceptional performance. Our rigorous evaluation spanning diverse deployment scenarios reveals that \disco{} delivers remarkable improvements, reducing both mean TTFT (6-78\%) and tail TTFT (11-52\%) while achieving cost savings of up to 83.6\%.

\subsection{Evaluation Setup}
\paragraph{Testbeds and Workloads.} 
Our testbed is a server with 4 NVIDIA A40 GPUs, each with 48GB of memory. 
We evaluate \disco{} using both commercial LLM traces and on-device deployments. For server-side evaluation, we collect traces from four production services: OpenAI's GPT-4o-mini \citep{gpt-4o-mini}, DeepSeek-V2.5 \citep{deepseek-v2_5}, Cohere's Command \citep{command}, and Hyperbolic-hosted LLaMA-3-70b-Instruct \citep{llama3-70b}. For on-device evaluation, we test three representative device-model configurations \citep{edgebenchmark}: Pixel 7 Pro with Bloom 1.1B (31.32/13.93 tokens/s for prefill/decode), Pixel 7 Pro with Bloom 560M (51.80/20.14 tokens/s), and Xiaomi 14 with Qwen 1.5 0.5B (79.90/21.47 tokens/s). These configurations span different compute-capability trade-offs in mobile environments. 
For end-to-end cost comparison, we quantify server costs using commercial API token pricing and device costs using FLOPs-based energy consumption. The detailed cost analysis can be found in Appendix~\ref{appendix:unified_cost}.

\paragraph{Baselines.} 
We compare \disco{} with four on-server, on-device, and cooperative deployments: 

\begin{denseitemize}
    \item \emph{vLLM}~\cite{vllm}: Processes all requests using remote server-based deployment. 
    
    \item \emph{llama.cpp}~\cite{llama_cpp}: Processes all requests using local device-based deployment.
    
    \item \emph{Stoch-S}: A server-constrained approach that randomly routes requests to the device while capping the server budget.
    
    \item \emph{Stoch-D}: A device-constrained approach that randomly routes requests to the server while capping the device budget.
\end{denseitemize}

For end-to-end cost comparison, we include two additional baselines: \emph{DiSCo-D w/o Migration} and \emph{DiSCo-S w/o Migration}.

\paragraph{Metrics.} 
We evaluate the system performance using both TTFT and TBT, including their mean and tail values. They are analyzed across varying cost budgets, defined as the ratio of input tokens processed by the constrained endpoint (device or server) to the total input tokens. For each experiment, we report the mean value over 10 runs. 

\subsection{End-to-end Performance}

\paragraph{\disco{} improves TTFT performance.} 

\begin{figure}[t]
    \centering
    \includegraphics[width=\columnwidth]{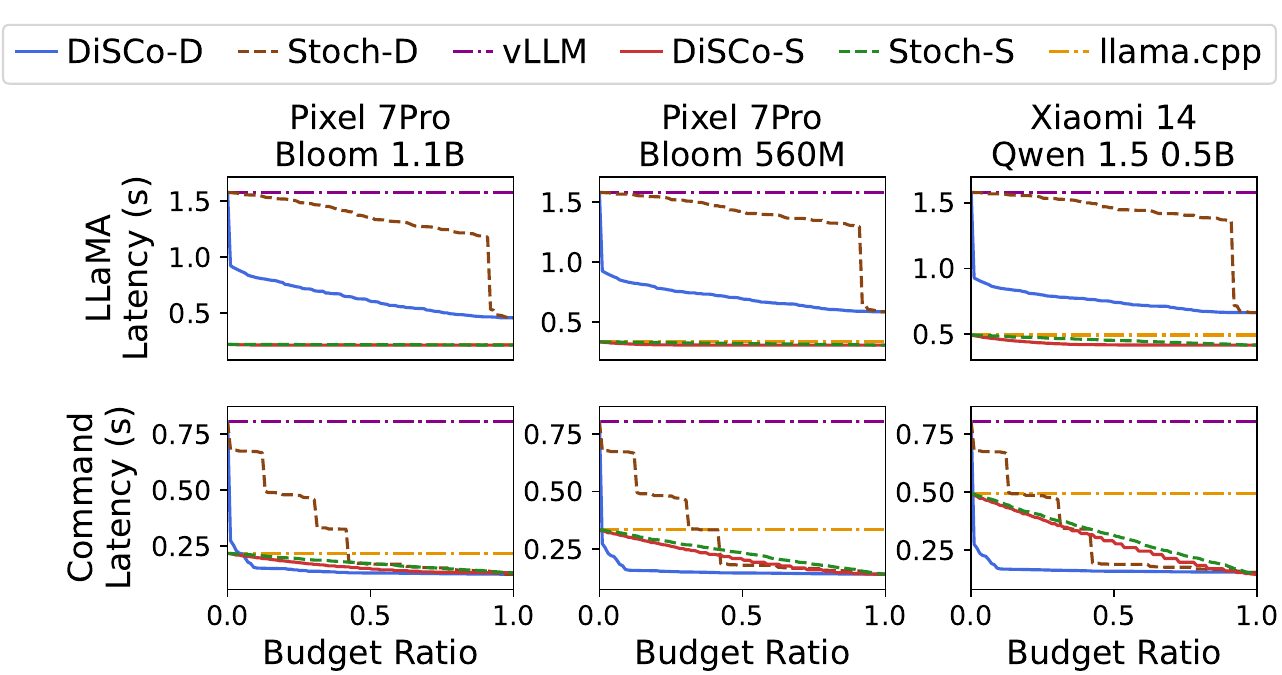}
    \vskip -0.1in
    \caption{Mean TTFT reduction of \disco{} remains significant on DiffusionDB trace.}
    \vskip -0.1in
    \label{fig:e2e-ttft-diffusiondb}
\end{figure}

\begin{figure*}[t]
    \centering
    \includegraphics[width=\textwidth]{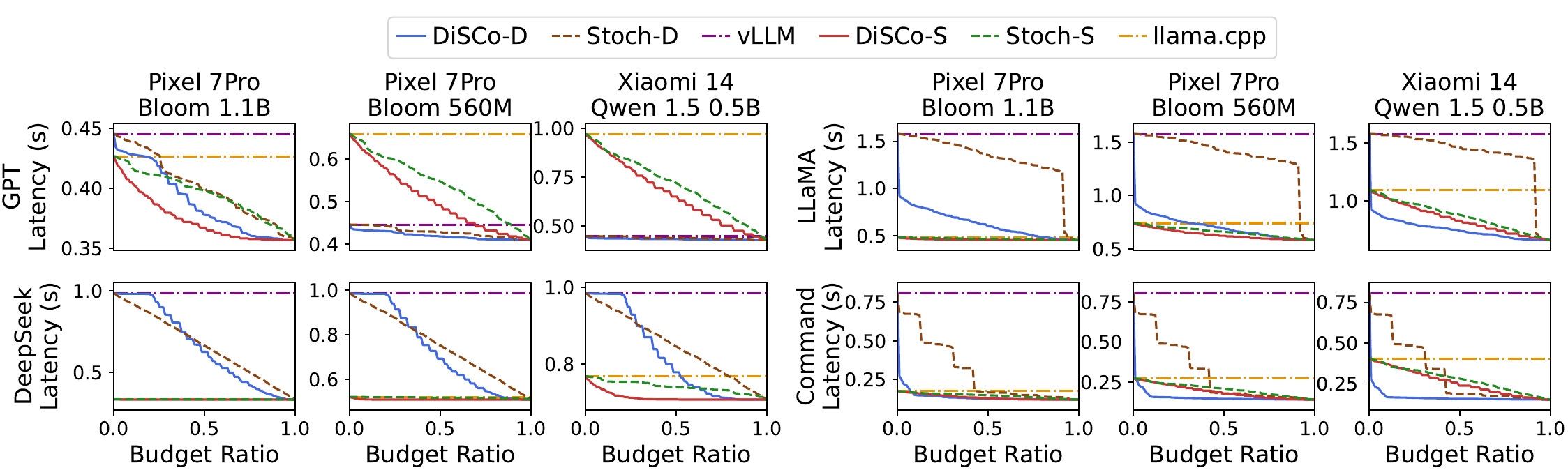}
    \vskip -0.1in
    \caption{Mean TTFT tested using four traces. \disco{} achieves superior TTFT performance than the baselines.
    }
    \vskip -0.1in
    \label{fig:e2e-ttft}
\end{figure*}

Figure~\ref{fig:e2e-ttft} and Table~\ref{tab:tail-ttft} show that \disco{} significantly outperforms baseline methods in both device- and server-constrained settings, showing improvements across mean and tail (P99) TTFT metrics for various services, including GPT, LLaMA, DeepSeek, and Command. In the GPT experiments, \disco{} demonstrates particularly notable tail latency reductions, decreasing P99 TTFT by up to 40\% relative to Stochastic dispatching across all device configurations, while mean TTFT is also reduced substantially, with reductions between 20-30\% across diverse budget ratios.
In the LLaMA setup, we observe a unique trade-off pattern. For budget ratios below 20\% when the device is the constrained endpoint, \disco{} exhibits a slightly higher mean TTFT than the baseline. This outcome is intentional, as \disco{} prioritizes tail latency reduction in low-budget scenarios, yielding substantial gains in P99 TTFT—reducing tail latency by up to 50\%. 

DeepSeek and Command experiments demonstrate similar patterns of improvement as the previous two traces, with \disco{} consistently outperforming baseline approaches. In the DeepSeek scenario, \disco{} maintains stable latency even as the budget ratio increases, whereas the baseline systems show increasing latency variance. 

 \begin{figure*}[t]
     \centering
     \includegraphics[width=\textwidth]{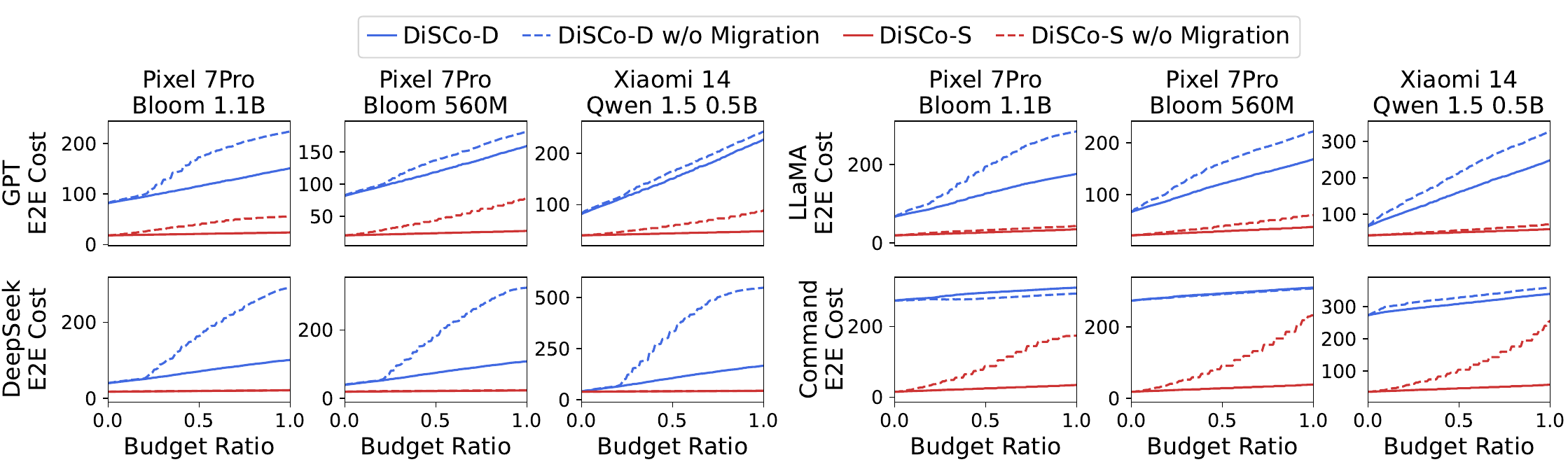}
     \vskip -0.1in
     \caption{The migration mechanism in \disco{} achieves superior end-to-end cost.
     }
     \vskip -0.1in
     \label{fig:e2e-cost}
\end{figure*}

\setlength{\tabcolsep}{1.5pt}
\begin{table}[t]
    \centering
    \footnotesize
    \begin{tabularx}{\linewidth}{ccccc}
    \toprule
    \multirow{3}{*}{\textbf{Platform}} & \multirow{3}{*}{\textbf{Constraint}} & \multicolumn{3}{c}{\textbf{Tail TTFT Reduction}} \\
    & & \textbf{\thead{Pixel 7Pro \\ B-1.1B}}  &  \textbf{\thead{Pixel 7Pro \\ B-560M}} & \textbf{\thead{Xiaomi 14 \\ Q-0.5B}} \\
    \midrule
    \multirow{2}{*}{GPT} & Server & 23.85\% & 37.41\% & 44.04\% \\
    & Device & 26.39\% & 21.48\% & 16.32\% \\
    \hline
    \multirow{2}{*}{LLaMA} & Server & 11.08\% & 23.09\% & 26.29\% \\
    & Device & 35.67\% & 29.30\% & 21.29\% \\
    \hline
    \multirow{2}{*}{DeepSeek} & Server & 0.00\%* & 3.88\% & 15.53\% \\
    & Device & 30.91\% & 28.01\% & 25.08\% \\
    \hline
    \multirow{2}{*}{Command} & Server & 47.93\% & 50.93\% & 52.23\% \\
    & Device & 34.78\% & 31.53\% & 24.42\% \\
    \bottomrule
    \end{tabularx}
    \caption{Average reduction of tail TTFT compared to stochastic dispatching across the whole cost budget range. Devices include Pixel 7 Pro and Xiaomi 14, while models include Bloom-1.1B, Bloom-560M, and Qwen-1.5-0.5B. (*Tail TTFT remains constant.)}
    \label{tab:tail-ttft}
\end{table}

\setlength{\tabcolsep}{4pt}
\begin{table}[t]
    \centering
    \footnotesize
    \begin{tabularx}{\linewidth}{ccccc}
    \toprule
    \multirow{2}{*}{\textbf{Trace}} & \multirow{2}{*}{\textbf{Constraint}} & \textbf{Mean} & \textbf{P99} & \textbf{TBT} \\
    & & \textbf{delay\_num} & \textbf{delay\_num} & \textbf{P99} \\
    \midrule
    \multirow{2}{*}{GPT} & Server & 4.21 & 9.40 & 0.209 \\
    & Device & 6.59 & 6.59 & 0.217 \\
    \hline
    \multirow{2}{*}{LLaMA} & Server & 5.53 & 11.00 & 0.209 \\
    & Device & 10.01 & 10.01 & 0.217 \\
    \hline
    \multirow{2}{*}{DeepSeek} & Server & 8.13 & 11.00 & 0.209 \\
    & Device & 17.17 & 17.17 & 0.217 \\
    \hline
    \multirow{2}{*}{Command} & Server & 3.25 & 8.00 & 0.209 \\
    & Device & 8.54 & 8.54 & 0.217 \\
    \bottomrule
    \end{tabularx}
    \caption{Performance metrics for different models under server and device constraints, showing the number of delayed tokens during migration and TBT (Time Between Tokens) P99 statistics. The average is computed over the requests that have performed the migration. 
    }
    \vspace{-.5cm}
    \label{tab:performance-metrics}
\end{table}

\paragraph{\disco{} retains TBT performance while lowering the cost.} 
Table~\ref{tab:performance-metrics} evaluates \disco{}'s TBT performance across various traces under both server and device constraints. For requests involving migration, we measure two key metrics: the average number of migrations per request and the tail (P99) TBT latency. Results show that while migrations delay only a negligible number of tokens compared to typical generation lengths of hundreds or thousands of tokens, they do not impact the perceived token delivery smoothness, demonstrating \disco{}'s ability to maintain consistent streaming performance even during endpoint transitions.

As shown in Figure~\ref{fig:e2e-cost}, our token-level migration mechanism substantially reduces the end-to-end cost across all evaluated scenarios. For device-constrained cases (\disco{}-D), the migration mechanism achieves up to 72.7\% cost reduction compared to the non-migration baseline, with the improvement being most significant at higher budget ratios. Similarly, in server-constrained scenarios (\disco{}-S), the cost reduction reaches 83.6\%, particularly evident in DeepSeek and Command model deployments. These significant cost reductions are consistently observed across device-model pairs.

\subsection{Performance Breakdown and Ablation Study}

\paragraph{Impact of Prompt Arrival Interval.} 
To evaluate our system under realistic workload patterns, we conduct experiments using stratified sampling based on request arrival rate from DiffusionDB \citep{diffusiondb}. Specifically, we select traces from ten users across different activity levels to capture diverse interaction patterns. We pair these real-world request intervals with prompts randomly drawn from the Alpaca dataset \citep{alpaca}. The results shown in Figure~\ref{fig:e2e-ttft-diffusiondb} demonstrate that \disco{}'s performance advantages persist across varied user activity patterns.

\begin{figure}[t]
    \centering
    \includegraphics[width=0.4\textwidth]{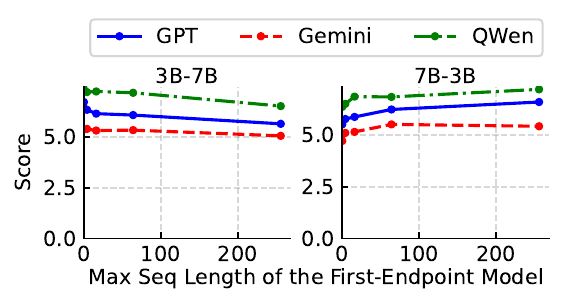}
    \vskip -0.1in
    \caption{Response quality evaluation. Each subplot represents a distinct model pair configuration (e.g., 3B-7B indicates migration from a 3B to a 7B model). The x-axis shows the maximum sequence length processed by the first endpoint before migration, while the y-axis shows the quality scores assigned by different LLM judges. Results demonstrate consistent quality preservation across various migration scenarios.}
    \vskip -0.1in
    \label{fig:accuracy_analysis}
\end{figure}

\paragraph{Quality of Generated Responses.}
We conduct comprehensive experiments using instruction-following tasks on multiple model configurations. We employ three LLM-based judges (GPT-4o, Gemini1.5-pro, and QWen2.5-72b) to assess the response quality, and examine two representative migration scenarios: from a smaller to larger model (3B-7B) and vice versa (7B-3B). Figure~\ref{fig:accuracy_analysis} shows that \disco{} maintains quality scores across different sequence lengths, migration patterns, and judges. Detailed results are presented in Appendix~\ref{appendix:accuracy-eval}.

\begin{figure}
    \subfigure[\disco{}-D \label{fig:overhead_device}]{\includegraphics[width=0.5\columnwidth]{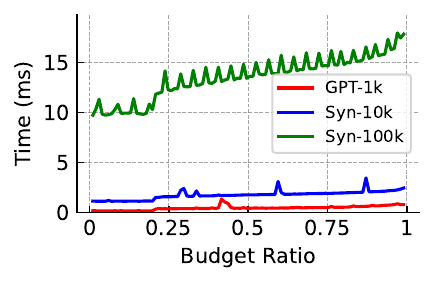}}\hfill
    \subfigure[\disco{}-S \label{fig:overhead_server}]{\includegraphics[width=0.5\columnwidth]{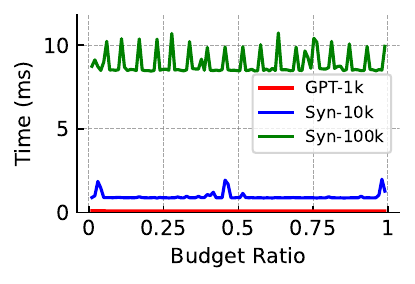}}
    \caption{\disco{}'s overhead is trivial and can scale well.}
    \vspace{-.4cm}
    \label{fig:overhead}
\end{figure}

\paragraph{Scalability Analysis.}
We conducted comprehensive performance evaluations of \disco{}-D and \disco{}-S on a MacBook Pro with M1 processor, using both synthetic datasets and a real-world GPT trace of 1,000 records, across target frequencies from 0 to 1. 
To generate synthetic data that accurately reflects real-world scenarios, we fitted log-normal distributions to the prompt lengths and TTFT from the real trace by following the mean and standard deviation of the logarithm. 
As shown in Figure~\ref{fig:overhead}, for \disco{}-S, the execution time showed remarkable efficiency: 0.128 ms for the real trace with 1K samples, scaling to just 0.969 ms and 9.082 ms for synthetic datasets of 10K and 100K samples, respectively. \disco{}-D, while being more computationally intensive, still maintained practical performance levels: 0.486 ms, 1.741 ms, and 14.856 ms for 1K, 10K, and 100K samples, respectively.

%% file: src/conclusion.tex
\section{Conclusions}

This paper introduces \disco{}, a device-server cooperative scheduler that addresses QoE and cost challenges in LLM serving for real-time conversational applications for end users. \disco{} uses cost-aware scheduling and token-level migration to dynamically optimize TTFT and TBT across device and server endpoints. Our evaluations on real-world traces from platforms like GPT and DeepSeek show that \disco{} significantly improves both TTFT and TBT while reducing costs.

%% file: src/appendix/related_work.tex
\section{Additional Related Work}\label{appendix:related_work}

\paragraph{General LLM Inference.}
\label{sec:general_llm_inference}
LLMs generate text responses auto-regressively, producing one token at a time based on preceding tokens. The process consists of two stages that can potentially be executed on different endpoints: (i) \emph{Prefill stage}: The model processes the input text (prompt), calculates and stores intermediate model states--i.e., the key and value cache (KV cache) of tokens--to generate the first token. A token represents a word or part of a word that the model can interpret. Once the first token is generated, it is appended to the end of the prompt and the generation process moves on to the (ii) \emph{Decode stage}: The model processes the updated prompt (including previously generated tokens) to generate the next token. The decode stage continues until a stopping condition is met (e.g., reaching an end-of-sequence token or the maximum generation length).

\paragraph{On-Server LLM Serving.}
Existing works have focused on GPU kernel optimization \citep{flashattention,flashinfer}, KV-cache management \citep{lin2024infinite,scissorhands,mooncake}, model parallelism \citep{shoeybi2019megatron,pope2023efficiently,liu2023ring}, quantization \citep{smoothquant,awq,dettmers2022gpt3}, and scheduling \citep{orca,vllm,sarathi,wang2025hygen}.
For example, Orca \cite{orca} introduced continuous batching to improve serving throughput, while vLLM \cite{vllm} developed PagedAttention to reduce LLM memory restraint. Sarathi \citep{sarathi} implemented chunked prefill to mitigate inter-request interference within batches. 
Andes \citep{andes} addresses QoE for individual requests from the server side but lacks awareness of network jitter and device potential. These server-side advancements complement \disco{}'s design.

\paragraph{On-Device LLMs.}
Google's Gemini Nano \cite{gemini_nano} and Apple's Apple Intelligence \cite{appleintelligence} have been integrated into Android OS and iOS devices, respectively.
MLC-LLM \cite{mlc-llm} and llama.cpp \citep{llama_cpp} efficiently deploy various LLMs on devices. PowerInfer \cite{powerinfer} and PowerInfer-2 \cite{powerinfer2} optimize on-device LLM inference by leveraging sparsity in model activations. \disco{} acts as a middle layer to schedule and migrate response generation between servers and devices.

%% file: src/appendix/cold_start.tex
\section{Cold Start Evaluation}\label{appendix:cold_start}

This section presents cold start performance measurements for the Qwen-2.5 model series across different hardware configurations.
The experiments were conducted on two platforms: Windows 10 with NVIDIA RTX 3060 12GB and Linux with NVIDIA A40 48GB. A fixed prompt "How to use GitHub?" was used throughout all experiments. We measured two critical metrics: model loading time and TTFT for Qwen-2.5 models ranging from 0.5B to 7B parameters, all using FP16 precision. The experimental setup consisted of 10 measurement runs, with two additional warmup runs to ensure measurement stability. It is worth noting that such warmups can potentially mask the true gap between model loading and prompt prefill time due to various optimizations, including OS page cache. To maintain authentic cold start conditions, we explicitly cleared the CUDA cache and performed garbage collection before each run.

The results revealed several significant patterns. On the RTX 3060, the loading time exhibits an approximately linear increase with model size, ranging from 1.29s for the 0.5B model to 4.45s for the 3B model. While TTFT follows a similar trend, the processing time is substantially lower, ranging from 0.051s to 0.145s. On the A40 GPU, despite observing longer loading times, TTFT is significantly reduced across all models, maintaining a remarkably consistent value regardless of model size. These findings indicate that while model loading remains more resource-intensive on our Linux setup, the inference performance benefits substantially from the A40's superior computational capabilities.

\begin{table}[ht]
    \centering
    \footnotesize
    \vskip 0.1in
    \begin{tabular}{p{2.5cm}cccc}
    \toprule
    \textbf{Metric} & \textbf{0.5B} & \textbf{1.5B} & \textbf{3B} & \textbf{7B} \\
    \midrule
    \multicolumn{5}{c}{\textbf{Windows 10 (NVIDIA RTX 3060 12GB)}} \\
    \midrule
    Load Time (s) & 1.29 & 2.48 & 4.45 & - \\
    TTFT (s) & 0.051 & 0.105 & 0.145 & - \\
    \midrule
    \multicolumn{5}{c}{\textbf{Linux (NVIDIA A40 48GB)}} \\
    \midrule
    Load Time (s) & 1.53 & 3.12 & 5.72 & 13.43 \\
    TTFT (s) & 0.025 & 0.026 & 0.033 & 0.033 \\
    \bottomrule
    \end{tabular}
    \caption{Model loading time during cold start can significantly slow down TTFT. Average Qwen-2.5 model performance over 10 runs. The 7B model exceeds the memory capacity of the RTX 3060 and thus cannot be evaluated.}
    \label{tab:cold-start}
\end{table}

%% file: src/appendix/prediction.tex
\section{Prediction-based Model Selection}\label{appendix:prediction}

This section provides a comparative analysis of several TTFT prediction methods. For selecting the endpoint with a lower TTFT for each request, TTFT prediction is imperative. For on-device inference, TTFT prediction is straightforward, as TTFT exhibits a linear relationship with prompt length. Conversely, on-server TTFT is characterized by high variability, posing challenges for prediction. Moreover, the prediction method itself must be computationally efficient, as its overhead also contributes to end-to-end TTFT.

Table~\ref{tab:model-comparison} presents a comparative analysis of four common lightweight time-series-based prediction methods applied to traces collected from three prevalent LLM services. Our correlation analysis (Table~\ref{tab:correlation-analysis}) reveals no significant correlation between prompt length and TTFT; thus, prompt length is omitted as a feature in these prediction methods. We demonstrate that none of these methods offers sufficient accuracy for TTFT prediction.

\begin{table}[t]
    \centering
    \footnotesize
    \begin{tabular}{p{3.5cm}cc}
    \toprule
    \textbf{Model} & \textbf{MAPE(\%)} & \textbf{MAE(s)} \\
    \midrule
    \multicolumn{3}{c}{\textbf{Command}} \\
    \midrule
    Moving Average & 39.40 & 0.0899 \\
    ExponentialSmoothing & 53.51 & 0.1047 \\
    Random Forest & 39.33 & 0.0966 \\
    XGBoost & 35.43 & 0.0905 \\
    \midrule
    \multicolumn{3}{c}{\textbf{DeepSeek-V2.5}} \\
    \midrule
    Moving Average & 27.80 & 0.3959 \\
    ExponentialSmoothing & 27.39 & 0.3771 \\
    Random Forest & 32.97 & 0.4745 \\
    XGBoost & 27.51 & 0.4001 \\
    \midrule
    \multicolumn{3}{c}{\textbf{GPT-4o-mini}} \\
    \midrule
    Moving Average & 24.55 & 0.0995 \\
    ExponentialSmoothing & 20.88 & 0.0844 \\
    Random Forest & 28.68 & 0.1128 \\
    XGBoost & 24.83 & 0.0997 \\
    \midrule
    \multicolumn{3}{c}{\textbf{LLaMA-3-70b-Instruct}} \\
    \midrule
    Moving Average & 42.18 & 0.3312 \\
    ExponentialSmoothing & 40.27 & 0.3154 \\
    Random Forest & 49.67 & 0.3875 \\
    XGBoost & 43.94 & 0.3451 \\
    \bottomrule
    \end{tabular}
    \caption{Comparative analysis of Moving Average, Exponential Smoothing, Random Forest, and XGBoost prediction models across Command, DeepSeek, GPT, and LLaMA model traces. Metrics include Mean Absolute Percentage Error (MAPE) and Mean Absolute Error (MAE).}
    \label{tab:model-comparison}
\end{table}

To address potential concerns about the robustness of our server-side TTFT prediction under varying workload conditions, we emphasize that our distribution-based prediction model is specifically designed to handle workload bursts and dynamic patterns. The statistical distribution we maintain inherently captures these patterns, allowing the model to automatically adapt to changing workloads based on historical data. Furthermore, our system architecture supports dynamic distribution updates through a lightweight server API, which can efficiently transmit real-time server workload information using just tens or hundreds of numbers. This design ensures that our prediction model remains responsive and accurate even during significant load spikes or ephemeral workloads.

%% file: src/appendix/response_quality.tex
\section{Response Quality}\label{appendix:accuracy-eval}

This section examines the quality of responses generated by \disco{}, with a particular focus on quality preservation during endpoint transitions. We first establish bounds on generation quality, then present our evaluation methodology, and finally demonstrate through extensive experiments that \disco{} maintains consistent quality across different model configurations and tasks.

\subsection{Quality Bounds}

A critical aspect of \disco{} is maintaining generation quality during endpoint transitions. We employ a systematic approach to quality preservation~\cite{diba2017weakly,gupta2022semi,chen2023frugalgpt}. Specifically, for endpoints A and B with quality metrics $Q_A$ and $Q_B$ (measured by LLM scores or ROUGE scores), we find that any migrated sequence M with quality $Q_M$ satisfies:

\begin{equation}
    \min(Q_A, Q_B) \leq Q_M \leq \max(Q_A, Q_B)
\end{equation}

This bound ensures that migration does not degrade quality beyond the capabilities of individual endpoints.

\subsection{Evaluation Methodology}

\paragraph{Evaluation Framework}
We establish a comprehensive assessment framework encompassing both automated metrics and LLM-based evaluation. Our framework evaluates two distinct tasks:

\begin{itemize}
    \item \textbf{Translation Quality}: We assess Chinese-to-English translation on 500 data items from Flores\_zho\_Hans-eng\_Latn dataset~\cite{nllb2022,goyal2022flores} using the ROUGE-1 metric.
    
    \item \textbf{Instruction Following}: We evaluate 500 data items from the Alpaca dataset~\cite{alpaca} using our structured prompt template, with quality assessment performed by multiple LLM judges: Gemini1.5-pro, GPT-4o, and QWen2.5-72b-instruct.
    
    \item \textbf{Text Summarization}: We evaluate 100 data items from the CNN/DM~\cite{nallapati2016abstractive} using our structured prompt template, with quality assessment performed by DeepSeek-V3-0324.
    
    \item \textbf{Story Writing}: We evaluate 100 data items from the WritingPrompts~\cite{fan2018hierarchical} using our structured prompt template, with quality assessment performed by DeepSeek-V3-0324.
\end{itemize}

These two tasks are popular on end-user devices. Understandably, for complex tasks such as advanced math reasoning, we notice DisCo can lead to accuracy drops compared to the on-server model due to the limited capability of the on-device models, yet still achieves better performance than the on-device counterpart. 

\paragraph{Experimental Setup}
We configure our experiments with:
\begin{itemize}
    \item A fixed maximum generation length of 256 tokens
    \item First endpoint's maximum generation length varied through different (e.g., [0, 4, 16, 64, 256] tokens). 
    \item Four model combinations: 0.5B-7B, 3B-7B, 7B-0.5B, and 7B-3B (prefix and suffix denote the model sizes of first and second endpoints, respectively)
\end{itemize}

The generation transitions to the second endpoint when the first endpoint reaches its length limit without producing an end-of-generation token, creating natural boundary conditions for analysis.

For instruction-following tasks, we employ the following structured evaluation template:

\begin{footnotesize}
\begin{verbatim}
JUDGE_PROMPT = """Strictly evaluate the 
quality of the following answer on a scale
of 1-10 (1 being the worst, 10 being the 
best). First briefly point out the problem
of the answer, then give a total rating in
the following format.

Question: {question}

Answer: {answer}

Evaluation: (your rationale for the rating, 
as a brief text)

Total rating: (your rating, as a number 
between 1 and 10)
"""
\end{verbatim}
\end{footnotesize}

\subsection{Results and Analysis}

\subsubsection{Quality Metrics}
Our comprehensive evaluation reveals that the combined sequences' quality consistently remains bounded between individual model performance levels.

\begin{figure}[t]
    \centering
    \includegraphics[width=0.45\textwidth]{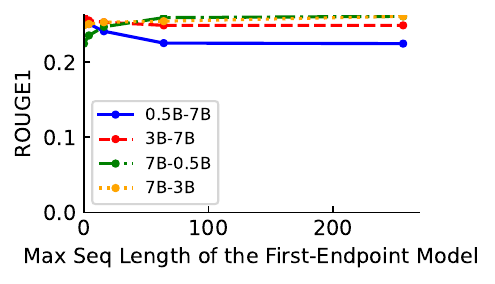}
    \\[12pt]
    \includegraphics[width=0.45\textwidth]{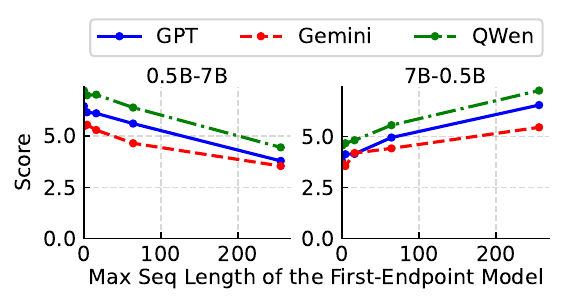}
    \caption{Quality evaluation results of \disco{}. The top figure shows translation quality evaluation using ROUGE-1 scores, demonstrating that \disco{} consistently achieves higher quality than the on-device baseline. The bottom figure presents evaluation scores from different LLM judges on instruction-following capabilities, where each subplot represents a different model pair comparison with varied first-endpoint model's maximum sequence length. The consistent patterns across different LLM judges demonstrate the robustness of our evaluation framework.}
    \label{fig:accuracy}
\end{figure}

\begin{figure}
    \subfigure[Summarization.]{\includegraphics[width=0.5\columnwidth]{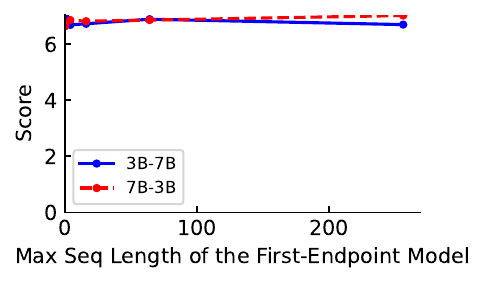}}\hfill
    \subfigure[Story Writing.]{\includegraphics[width=0.5\columnwidth]{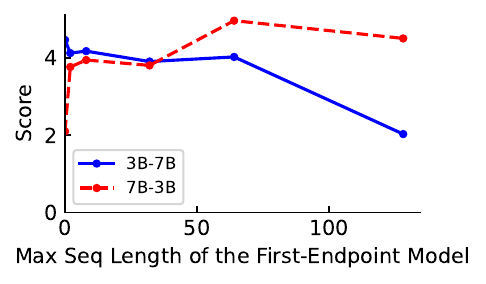}}
    \caption{Quality evaluation results with Summarization and Story Writing.}
    \label{fig:accuracy2}
\end{figure}

%% file: src/appendix/unified_cost.tex
\section{Experiment Settings for End-to-end Cost}
\label{appendix:unified_cost}

For on-device LLMs, we quantify cost using FLOPs (floating-point operations). For on-server LLM services, we use their respective pricing rates at the time of experimentation. We set the energy-to-monetary conversion ratio (\textit{energy\_to\_money}) to 0.3 \$ per million FLOPs for server-constrained experiments and 5 \$ per million FLOPs for device-constrained experiments. To establish a comprehensive cost model that enables direct comparison between device and server computation costs, we analyze both the computational complexity of on-device models through detailed FLOPs calculations (Section~\ref{sec:flops_analysis}) and the pricing structures of commercial LLM services (Section~\ref{appendix:llm_pricing}). The generation length limit is set to 128.

\subsection{Theoretical Cost Modeling}
\label{sec:cost_modeling}

Our unified cost model combines heterogeneous costs (server monetary and device energy) through a dynamic exchange rate $\lambda$, which users adjust based on preferences (e.g., battery level, server budget). We formalize this as:

\begin{equation}
    \text{Total Cost} = \underbrace{c_s^p \cdot l_s + c_s^d \cdot l_s}_{\text{Server Cost}} + \lambda \cdot \underbrace{\left(c_d^p \cdot l_d + c_d^d \cdot l_d\right)}_{\text{Device Cost}}
\end{equation}

where $l_s, l_d$ denote tokens processed on the server and the device, respectively. The optimization goal is to minimize $\mathbb{E}[\text{Total Cost}]$ while satisfying TTFT/TBT constraints. This formulation allows us to:

\begin{denseitemize}
    \item Balance between server monetary costs and device energy consumption
    \item Adapt to user preferences through the exchange rate $\lambda$
    \item Account for different costs in prefill ($c^p$) and decode ($c^d$) phases
    \item Optimize token allocation between server and device execution
\end{denseitemize}

The model's flexibility enables it to handle both device-constrained scenarios (where energy consumption is the primary bottleneck) and server-constrained scenarios (where API monetary costs dominate).

While we acknowledge that real-world device energy consumption can be influenced by factors such as battery levels, thermal throttling, and concurrency, we use the linear FLOPs-based measure as a generalizable proxy metric for two key reasons: (1) to cover energy consumption across diverse devices (phones, laptops, and edge servers) and models with varying architectures and quantizations that may exhibit different consumption patterns, and (2) to target short conversational tasks where on-device LLMs demonstrate sufficient efficiency and effectiveness. Importantly, since our dispatching and migration algorithms are both length-threshold-based, they can adapt to any complex energy consumption patterns as long as these patterns are predictable and guaranteed to consume more energy when prefilling or decoding additional tokens.

\subsection{FLOPs of On-Device LLMs}
\label{sec:flops_analysis}

To accurately quantify the computational cost per token in both prefill and decode stages, we conduct a detailed FLOPs analysis using three representative models: BLOOM-1.1B, BLOOM-560M, and Qwen1.5-0.5B. All models share a 24-layer architecture but differ in other parameters: BLOOM-1.1B ($d_{\text{model}}=1024$, 16 heads, FFN dim=4096), BLOOM-560M ($d_{\text{model}}=512$, 8 heads, FFN dim=2048), and Qwen1.5-0.5B ($d_{\text{model}}=768$, 12 heads, FFN dim=2048).

\paragraph{Per-token FLOPs computation.}
The total FLOPs for processing each token consist of five components:
\begin{align}
    \text{FLOPs}_{\text{total}} &= \text{FLOPs}_{\text{attn}} + \text{FLOPs}_{\text{ffn}} \nonumber \\
    &\quad+ \text{FLOPs}_{\text{ln}} + \text{FLOPs}_{\text{emb}} + \text{FLOPs}_{\text{out}}
\end{align}

For a sequence of length $L$, the attention computation differs between stages. In prefill:
\begin{align}
    \text{FLOPs}_{\text{attn}} &= n_{\text{layers}} \cdot \Big(3d_{\text{model}}^2 + \frac{L^2 d_{\text{model}}}{n_{\text{heads}}} \nonumber \\
    &\quad+ L d_{\text{model}} + d_{\text{model}}^2\Big)
\end{align}

While in decode, KV caching eliminates the quadratic term:
\begin{align}
    \text{FLOPs}_{\text{attn}} &= n_{\text{layers}} \cdot \Big(3d_{\text{model}}^2 + \frac{L d_{\text{model}}}{n_{\text{heads}}} \nonumber \\
    &\quad+ L d_{\text{model}} + d_{\text{model}}^2\Big)
\end{align}

Table~\ref{tab:prefill-decode} presents the total FLOPs across different sequence lengths. The decode phase maintains constant FLOPs regardless of sequence length due to KV caching, while prefill phase FLOPs increase with sequence length. A breakdown of computational cost by component (Table~\ref{tab:components}) reveals that embedding and output projection operations account for the majority of FLOPs, particularly in models with large vocabularies.

\begin{table}[t]
    \centering
    \footnotesize
    \begin{tabularx}{\linewidth}{lccc}
    \toprule
    \textbf{Length} & \textbf{BLOOM-1.1B} & \textbf{BLOOM-560M} & \textbf{Qwen-0.5B} \\
    \midrule
    \multicolumn{4}{l}{\textit{Prefill Phase}} \\
    L = 32 & 0.85 & 0.45 & 0.39 \\
    L = 64 & 0.93 & 0.50 & 0.45 \\
    L = 128 & 1.25 & 0.65 & 0.69 \\
    \midrule
    \multicolumn{4}{l}{\textit{Decode Phase}} \\
    L = 32 & 0.82 & 0.42 & 0.37 \\
    L = 64 & 0.82 & 0.42 & 0.37 \\
    L = 128 & 0.82 & 0.42 & 0.37 \\
    \bottomrule
    \end{tabularx}
    \caption{Prefill and Decode FLOPs (billions)}
    \vskip -0.1in
    \label{tab:prefill-decode}
\end{table}

\begin{table}[t]
    \centering
    \footnotesize
    \begin{tabularx}{\linewidth}{lccc}
    \toprule
    \textbf{Component} & \textbf{BLOOM-1.1B} & \textbf{BLOOM-560M} & \textbf{Qwen-0.5B} \\
    \midrule
    Embedding & 31.24 & 25.00 & 31.51 \\
    Attention & 13.01 & 10.00 & 16.56 \\
    FFN & 24.48 & 20.00 & 20.38 \\
    LayerNorm & 0.02 & 0.02 & 0.04 \\
    Output & 31.24 & 25.00 & 31.51 \\
    \bottomrule
    \end{tabularx}
    \caption{Component Ratios at L=128 (\%)}
    \vskip -0.1in
    \label{tab:components}
\end{table}

\subsection{LLM Service Pricing}\label{appendix:llm_pricing}
This section provides further details on the pricing of LLM services. Table \ref{tab:model-pricing} presents the pricing models for several commercial Large Language Models (LLMs) as of October 28, 2024. The pricing structure follows a dual-rate model, differentiating between input (prompt) and output (generation) tokens. These rates represent the public pricing tiers available to general users, excluding any enterprise-specific arrangements or volume-based discounts.

\begin{table}[t]
    \centering
    \footnotesize
    \begin{tabularx}{\linewidth}{lcccc}
    \toprule
    \textbf{Model} & \textbf{Vendor} & \textbf{Input price} & \textbf{Output price} \\
    \midrule
    DeepSeek-V2.5     & DeepSeek   & 0.14 & 0.28 \\
    GPT-4o-mini       & OpenAI     & 0.15 & 0.60 \\
    LLaMa-3.1-70b     & Hyperbolic & 0.40 & 0.40 \\
    LLaMa-3.1-70b     & Amazon     & 0.99 & 0.99 \\
    Command           & Cohere     & 1.25 & 2.00 \\
    GPT-4o            & OpenAI     & 2.50 & 10.0 \\
    Claude-3.5-Sonnet & Anthropic  & 3.00 & 15.0 \\
    o1-preview        & OpenAI     & 15.0 & 60.0 \\
    \bottomrule
    \end{tabularx}
    \caption{LLM service pricing (USD per 1M Tokens). Input prices refer to tokens in the prompt, while output prices apply to generated tokens.}
    \label{tab:model-pricing}
\end{table}

%% file: src/appendix/pseudocode.tex
\section{Pseudocode for Cost-Aware Adaptive Request Scheduling}\label{appendix:pseudocode_scheduling}

The request scheduling algorithm consists of three key components. Algorithm~\ref{alg:definitions} defines the input parameters and determines whether the scenario is device-constrained or server-constrained based on the relative costs. For device-constrained scenarios, Algorithm~\ref{alg:device_constrained} implements a wait-time strategy to protect tail latency while conserving device energy when possible. For server-constrained scenarios, Algorithm~\ref{alg:server_constrained} employs a length-based routing approach to optimize TTFT while maintaining the server budget constraint. These algorithms work together to achieve the dual objectives of minimizing latency and managing costs.

\begin{algorithm}[ht]
\caption{Variable Definitions and Constraints}
\label{alg:definitions}
\begin{algorithmic}[1]
\REQUIRE
    \STATE $p(l)$: Length distribution
    \STATE $F(t)$: TTFT CDF of server
    \STATE $b \in [0,1]$: Budget ratio 
    \STATE $c^p_d,c^d_d$: Device prefill/decode costs
    \STATE $c^p_s,c^d_s$: Server prefill/decode costs
    \STATE $\alpha \in (0,1)$: Tail ratio
\ENSURE Policy type based on cost constraints
\STATE \textbf{if} $\min(c^p_d,c^d_d) > \max(c^p_s,c^d_s)$ \textbf{then} Device-constrained
\STATE \textbf{else} Server-constrained
\end{algorithmic}
\end{algorithm}

\begin{algorithm}[ht]
\caption{Device-constrained Scheduling}
\label{alg:device_constrained}
\begin{algorithmic}[1]
\REQUIRE Variables from Algorithm \ref{alg:definitions}
\STATE // Phase 1: Set maximum wait time for tail protection
\STATE $w_{tail} \leftarrow F^{-1}(1 - \min(\alpha, b))$ 
\STATE // Initialize wait times for all prompt lengths
\STATE $W \leftarrow \{l: w_{tail} \text{ for all } l\}$

\IF{$b \leq \alpha$} 
    \RETURN $W$ \COMMENT{Use max wait time for all lengths}
\ENDIF

\STATE // Phase 2: Optimize wait times with remaining budget
\STATE available\_budget $\leftarrow b - \alpha$
\FOR{$l \in$ sort(support($p(l)$))}
    \STATE length\_cost $\leftarrow p(l) \cdot l \cdot (1-\alpha)$ 
    \IF{available\_budget $\geq$ length\_cost}
        \STATE $W[l] \leftarrow 0$ \COMMENT{Start device immediately}
        \STATE available\_budget $\leftarrow$ available\_budget - length\_cost
    \ELSE
        \STATE // Find optimal wait time that meets budget
        \STATE Find $w^* \in [0, w_{tail}]$ where:
        \STATE $F(w^*) \cdot$ length\_cost + (b - available\_budget) = $b$
        \STATE $W[l] \leftarrow w^*$
        \STATE \textbf{break}
    \ENDIF
\ENDFOR
\RETURN $W$ \COMMENT{Map from prompt lengths to wait times}
\end{algorithmic}
\end{algorithm}

\begin{algorithm}[ht]
\caption{Server-constrained Scheduling}
\label{alg:server_constrained}
\begin{algorithmic}[1]
\REQUIRE Variables from Algorithm \ref{alg:definitions}
\STATE // Find length threshold to split execution modes
\STATE Compute $l_{th}$ where: $\int_0^{l_{th}} l \cdot p(l) dl = (1-b) \cdot \int_0^\infty l \cdot p(l) dl$

\STATE // Initialize execution policy map
\STATE $P \leftarrow \emptyset$ 
\FOR{$l \in$ support($p(l)$)}
    \IF{$l < l_{th}$}
        \STATE $P[l] \leftarrow$ (1, 0) \COMMENT{$(I_d,I_s)$: Device only}
    \ELSE
        \STATE $P[l] \leftarrow$ (1, 1) \COMMENT{$(I_d,I_s)$: Concurrent execution}
    \ENDIF
\ENDFOR
\RETURN $P$ \COMMENT{Map from lengths to execution indicators}
\end{algorithmic}
\end{algorithm}